\def\paperlanguage{}
\pdfoutput=1 

\documentclass[letterpaper, 10 pt, conference]{ieeeconf}  

\usepackage{bm}
\usepackage{cite}
\usepackage{multirow}
\include{preamble}

\newcommand{\ctext}[1]{\raise0.2ex\hbox{\textcircled{\scriptsize{#1}}}}

\IEEEoverridecommandlockouts                              

\title{\LARGE \textbf
  {
    \switchlanguage%
    {%
      MEVIUS2: Practical Open-Source Quadruped Robot with\\Sheet Metal Welding and Multimodal Perception
    }%
    {%
      MEVIUS2: 板金溶接とマルチモーダル認識による\\実用的なオープンソース四脚歩行ロボット
    }%
  }
}

\author{Kento Kawaharazuka$^{1}$, Keita Yoneda$^{1}$, Shintaro Inoue$^{1}$, Temma Suzuki$^{1}$, Jun Oda$^{1}$, and Kei Okada$^{1}$
  \thanks{$^{1}$ The authors are with the Department of Mechano-Informatics, Graduate School of Information Science and Technology, The University of Tokyo, 7-3-1 Hongo, Bunkyo-ku, Tokyo, 113-8656, Japan.
    {\texttt\small [kawaharazuka, yoneda, s-inoue, t-suzuki, oda, k-okada]@jsk.imi.i.u-tokyo.ac.jp}
  }%
}

\begin{document}

\maketitle
\thispagestyle{empty}
\pagestyle{empty}

\begin{abstract}
  \switchlanguage%
  {%
    Various quadruped robots have been developed to date, and thanks to reinforcement learning, they are now capable of traversing diverse types of rough terrain.
    In parallel, there is a growing trend of releasing these robot designs as open-source, enabling researchers to freely build and modify robots themselves.
    However, most existing open-source quadruped robots have been designed with 3D printing in mind, resulting in structurally fragile systems that do not scale well in size, leading to the construction of relatively small robots.
    Although a few open-source quadruped robots constructed with metal components exist, they still tend to be small in size and lack multimodal sensors for perception, making them less practical.
    In this study, we developed MEVIUS2, an open-source quadruped robot with a size comparable to Boston Dynamics' Spot, whose structural components can all be ordered through e-commerce services.
    By leveraging sheet metal welding and metal machining, we achieved a large, highly durable body structure while reducing the number of individual parts.
    Furthermore, by integrating sensors such as LiDARs and a high dynamic range camera, the robot is capable of detailed perception of its surroundings, making it more practical than previous open-source quadruped robots.
    We experimentally validated that MEVIUS2 can traverse various types of rough terrain and demonstrated its environmental perception capabilities.
    All hardware, software, and training environments can be obtained from Supplementary Materials or \href{https://github.com/haraduka/mevius2}{\textcolor{magenta}{github.com/haraduka/mevius2}}.
  }%
  {%
    これまで様々な四足歩行ロボットが開発されてきており, 強化学習により多様な不整地を踏破可能になってきた.
    また, それらの設計をオープンソースとして公開し, 研究者自らが自由にロボットを構築・改変することができる流れが出来つつある.
    一方で, これまでのオープンソース四脚ロボットは主に3Dプリンタによる構築を前提としており, 構造的に脆く, 大きさもスケールしないため比較的小さなロボットが構築されてきた.
    一部金属により構成されるオープンソース四脚ロボットが存在するが, 未だサイズとして小さく, 認識のためのマルチモーダルなセンサもないため, 実用的とは言い難い.
    そこで本研究では, 全ての構造部品をE-Commerceにより発注可能かつ, Boston Dynamics Spotと同等のサイズを持つオープンソース四脚ロボットMEVIUS2を開発した.
    板金溶接と金属切削を駆使することで, 部品点数を削減しつつ, 非常に大きな身体構造を高い強度を持って構築することができる.
    また, LiDARや高いダイナミックレンジを持つカメラを統合することで, 周囲の情報を細かく認識し, これまでのオープンソース四脚ロボットに比べてより実用的な構成としている.
    MEVIUS2が多様な不整地を構築可能であること, その環境認識能力について実験を通して確認した.
    本研究の全てのハードウェア・ソフトウェア・学習環境はSupplementary Materialsに含まれている (採択後, GitHubに公開する予定である).
  }%
\end{abstract}

\section{INTRODUCTION}\label{sec:introduction}
\switchlanguage%
{%
  A wide variety of quadruped robots have been developed to date \cite{fujita1998quadruped, katz2019minicheetah, hutter2016anymal}, and their advancement has been remarkable.
  This progress has been greatly driven by the emergence of compact, low-reduction, high-torque motors developed in the MIT Cheetah series \cite{bosworth2015superminicheetah, katz2019minicheetah, bledt2018cheetah3}.
  Combined with advances in reinforcement learning, these developments have significantly enhanced the locomotion capabilities of quadruped robots \cite{hwangbo2019anymal, lee2020anymal, miki2022anymal}.
  Today, various quadruped robots such as ANYMAL \cite{hutter2016anymal}, Unitree Go1 \cite{unitree2022go1}, and DEEPRobotics Lite3 \cite{deeprobotics2023lite3} are commercially available.
  However, these commercial robots have several limitations: low-level control is often inaccessible, the design cannot be modified to suit specific research purposes, and users cannot easily repair the hardware themselves.
  If research institutions and individual researchers could engage with all aspects of development -- from mechanical design to low-level control -- it would greatly expand the possibilities for research.
}%
{%
  これまで様々な四脚ロボットが開発されてきており\cite{fujita1998quadruped, katz2019minicheetah, hutter2016anymal}, その発展には目覚ましいものがある.
  この流れは, MIT Cheetah シリーズ\cite{bosworth2015superminicheetah, katz2019minicheetah, bledt2018cheetah3}において開発された扁平型の低減速比で高トルクなモータの登場によるところが大きい.
  そして, これらは強化学習の発展と合わさり, 四脚ロボットに高い歩行能力を付与するに至っている\cite{hwangbo2019anymal, lee2020anymal, miki2022anymal}.
  現在は, ANYMAL \cite{hutter2016anymal}やUnitree Go1 \cite{unitree2022go1}, DEEPRobotics Lite3 \cite{deeprobotics2023lite3}をはじめとする様々な四脚ロボットが購入可能となっている.
  一方で, これらのロボットは低レイヤの制御が触りにくい, 自身の用途に合わせて設計を変更できない, 壊れても自身で修正できないなどの問題がある.
  研究機関・研究者個人が設計から低レイヤ制御まで含めて全てに触れることが可能になれば, 研究の幅は大きく広がると考えられる.
}%

\begin{figure}[t]
  \centering
  \includegraphics[width=0.95\columnwidth]{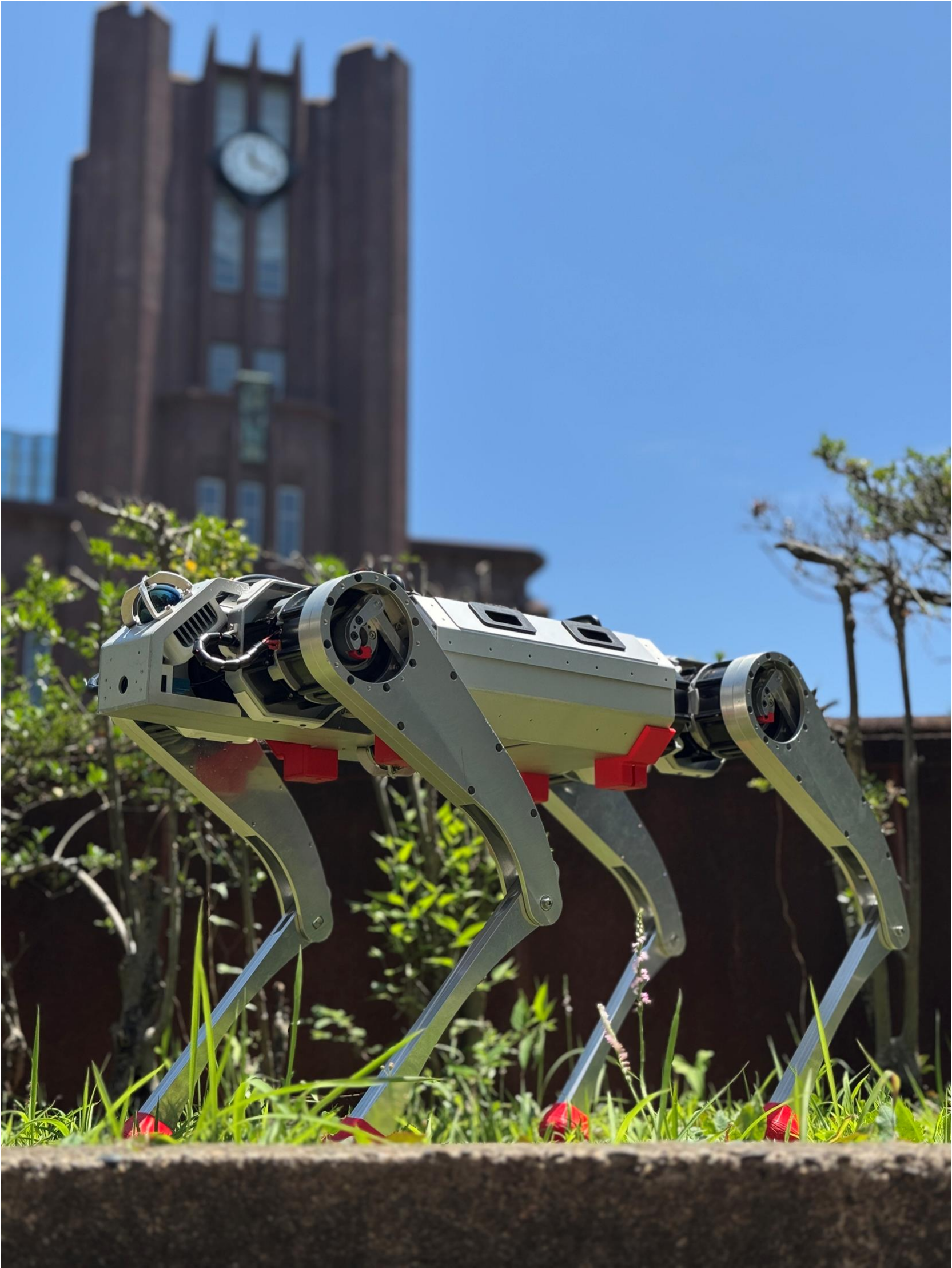}
  \vspace{-1.0ex}
  \caption{MEVIUS2 -- Practical open-source quadruped robot with sheet metal welding and multimodal perception, developed in this study.}
  \label{figure:mevius2}
  \vspace{-3.0ex}
\end{figure}

\begin{figure*}[t]
  \centering
  \includegraphics[width=1.95\columnwidth]{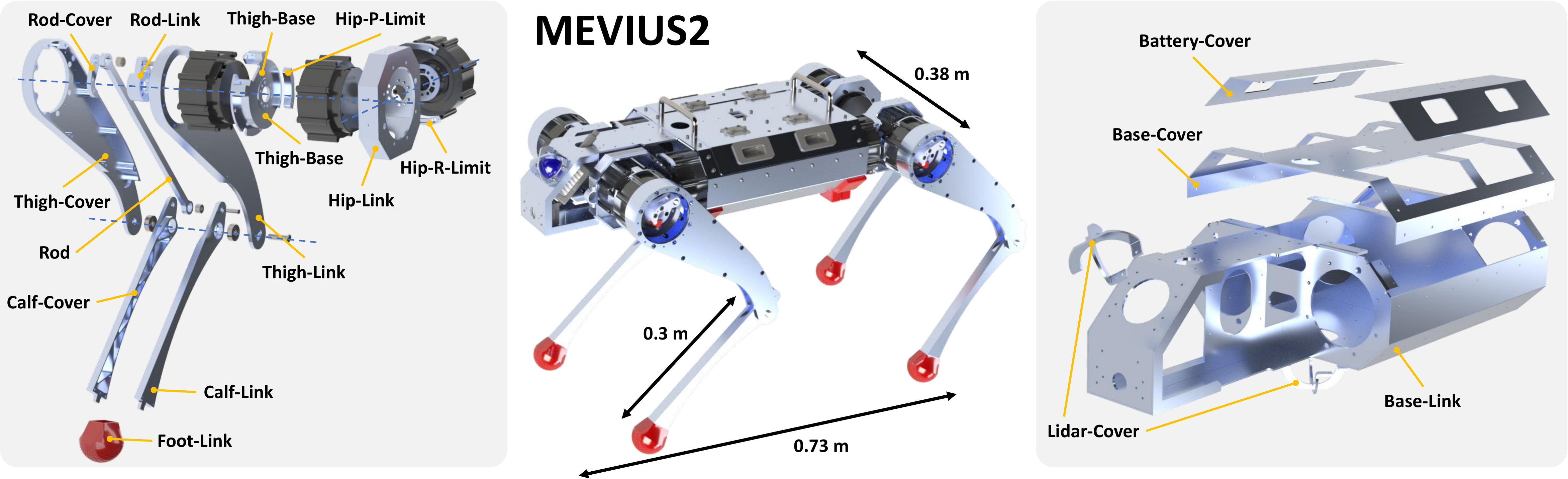}
  \vspace{-1.0ex}
  \caption{Design overview of MEVIUS2: excluding mirrored parts, the robot consists of 16 unique metal components, three of which are fabricated using sheet metal welding to achieve complex and large geometries as single integrated parts.}
  \label{figure:mevius2-design}
  \vspace{-2.0ex}
\end{figure*}

\switchlanguage%
{%
  Building on this trend, several open-source quadruped robots have recently been developed, with both hardware and software fully released to the public.
  Notable examples include Solo \cite{grimminger2020solo, leziart2021solo12} and PAWDQ \cite{joonyoung2021pawdq}.
  These robots can be built entirely from scratch by anyone using commercially available motors, circuits, and 3D printing.
  Thanks to the flexibility of 3D printing and the ease with which individual researchers can produce structural parts, many open-source quadruped robots have been developed using plastic components fabricated by 3D printers \cite{grimminger2020solo, leziart2021solo12, joonyoung2021pawdq, kau2021stanfordpupper, kau2019stanforddoggo, garc2020charlotte, sprowtz2018oncilla, lohmann2012aracna, rahme2021quadruped, xrobots2018opendog, quadvm2024tinymal}.
  However, 3D-printed structures tend to be mechanically fragile and do not scale well in size.
  As a result, most open-source quadruped robots to date have been relatively small, and few examples exist of such robots traversing diverse outdoor terrains.
  While a few open-source quadruped robots made with metal components have been developed \cite{kawaharazuka2024mevius}, their size has remained comparable to that of the Unitree Go1 \cite{unitree2022go1}.
  With such configurations, tasks like walking on stairs that are commonly used by humans remain challenging.
  Moreover, existing open-source quadruped platforms generally lack multimodal perception systems with LiDARs and cameras.
  Solving these issues would allow anyone to build a more practical quadruped robot from the ground up.

  To address this, we developed MEVIUS2, a metal-based open-source quadruped robot that matches the size of Boston Dynamics' Spot \cite{bostondynamics2025spot} and can be assembled entirely from components ordered via e-commerce.
  By leveraging sheet metal welding, we achieved a large and robust structure while significantly reducing the number of parts.
  All machined and welded metal components can be automatically quoted and ordered from STEP files or part numbers using meviy \cite{misumi2024meviy}, a machining and fabrication service by MISUMI.
  MEVIUS2 is also equipped with multimodal perception using LiDARs and a high dynamic range camera, enabling detailed sensing of its environment.
  We applied reinforcement learning to the developed MEVIUS2 and confirmed through experiments that it can successfully traverse various rough terrains.
  By releasing all hardware, software, and training environments, we aim to empower research institutions and individual researchers to modify and extend the platform -- facilitating the creation of even more innovative research outcomes.
}%
{%
  この流れを受けて, 現在ではハードウェアからソフトウェアまで全てを公開したオープンソース四脚ロボットが複数開発されている.
  その代表例としてSolo \cite{grimminger2020solo, leziart2021solo12}やPAWDQ \cite{joonyoung2021pawdq}が挙げられる.
  これらのロボットは既成品のモータや回路, 3Dプリンタを駆使することで, 誰でも一から構築することができるようになっている.
  3Dプリントは自由形状が容易かつ, 研究者個人でも比較的簡単に構造部品を調達できることから, 多くの3Dプリンタによるプラスチック製の四脚ロボットが開発されてきた\cite{grimminger2020solo, leziart2021solo12, joonyoung2021pawdq, kau2021stanfordpupper, kau2019stanforddoggo, garc2020charlotte, sprowtz2018oncilla, lohmann2012aracna, rahme2021quadruped}.
  その一方で, 3Dプリンタによる造形は構造的に脆く, ロボットの大きさがスケールしないという問題がある.
  そのため, これまで開発されてきたオープンソース四脚ロボットはどれも比較的サイズが小さく, 屋外環境で様々な不整地を歩行している例は極めて少ない.
  金属により構成されたオープンソース四脚ロボットが一部開発されているが\cite{kawaharazuka2024mevius}, 未だそのサイズはUnitree Go1 \cite{unitree2022go1}と同等の域を出ない.
  これでは, 人間が一般的に使用する階段の歩行などは難しい.
  また, これまでのオープンソース四脚ロボットにはLiDARやカメラに基づくマルチモーダルな認識が備わっていない.
  これらの問題を解決することができれば, より実用的な四脚ロボットを誰でも一から構築することが可能となるだろう.

  そこで本研究では, 全ての部品をE-Commerceにより調達可能かつ, Boston Dynamics Spot \cite{bostondynamics2025spot}と同等なサイズを持つ金属製のオープンソース四脚ロボットMEVIUS2を開発した.
  板金溶接を駆使することで, 非常に大きな構造を, 強度を保ちつつ構築することができ, ロボットの部品点数を大幅に削減することができる.
  全ての金属切削部品と板金溶接部品はMISUMIの切削・加工サービスであるmeviy \cite{misumi2024meviy}によりSTEPファイルまたは型番から自動的に見積り・発注することが可能である.
  また, LiDARとハイダイナミックレンジカメラによるマルチモーダル認識を備え, 環境情報を詳細に取得することを可能とした.
  開発したMEVIUS2に強化学習を適用し, 実験を通して様々な不整地を踏破できることを確認した.
  全てのハードウェア・ソフトウェア・学習環境を公開することで, 研究機関や研究者個人がそれを修正・改変し, より面白い研究成果が出てくることを期待する.
}%

\begin{figure}[t]
  \centering
  \includegraphics[width=0.9\columnwidth]{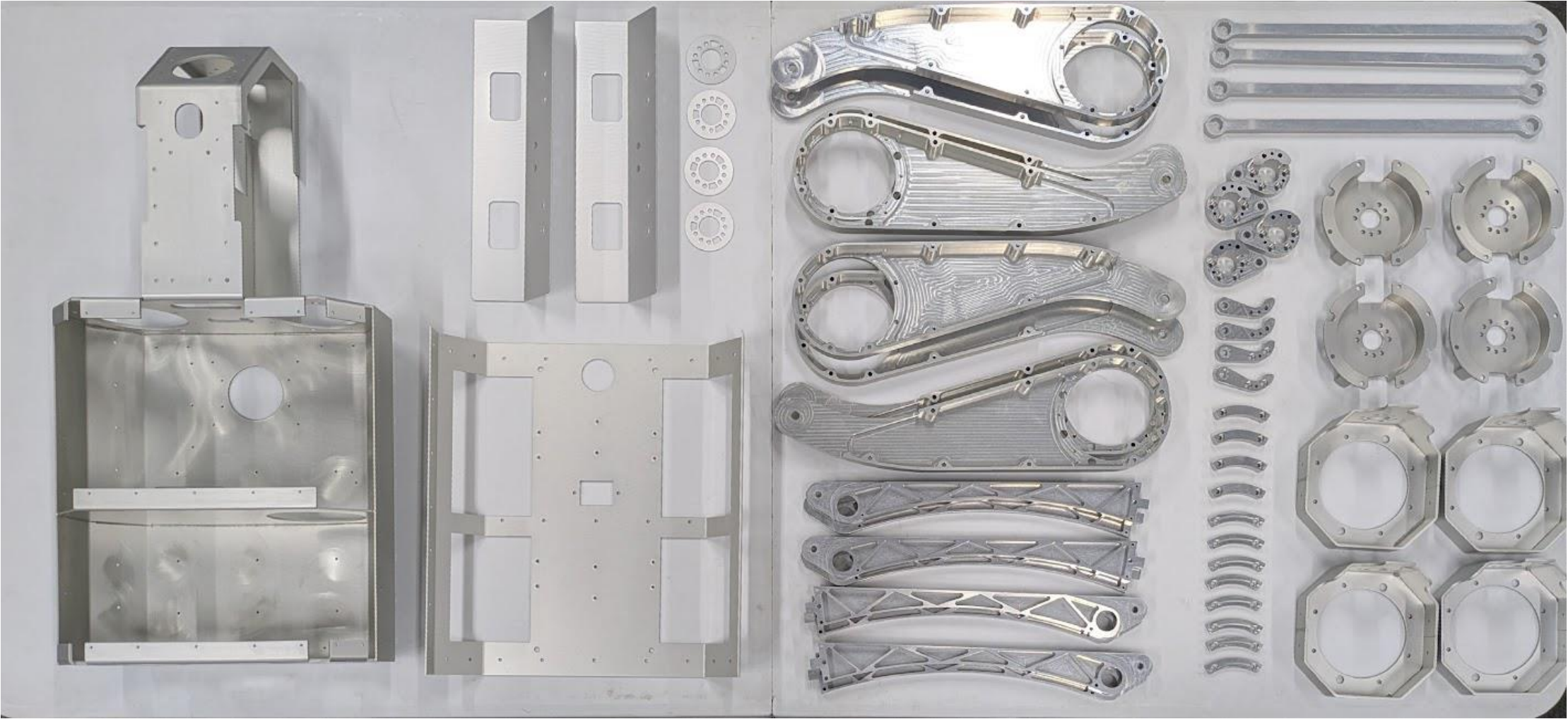}
  \vspace{-1.0ex}
  \caption{All metal components used in MEVIUS2.}
  \label{figure:mevius2-parts}
  \vspace{-3.0ex}
\end{figure}

\begin{figure}[t]
  \centering
  \includegraphics[width=0.85\columnwidth]{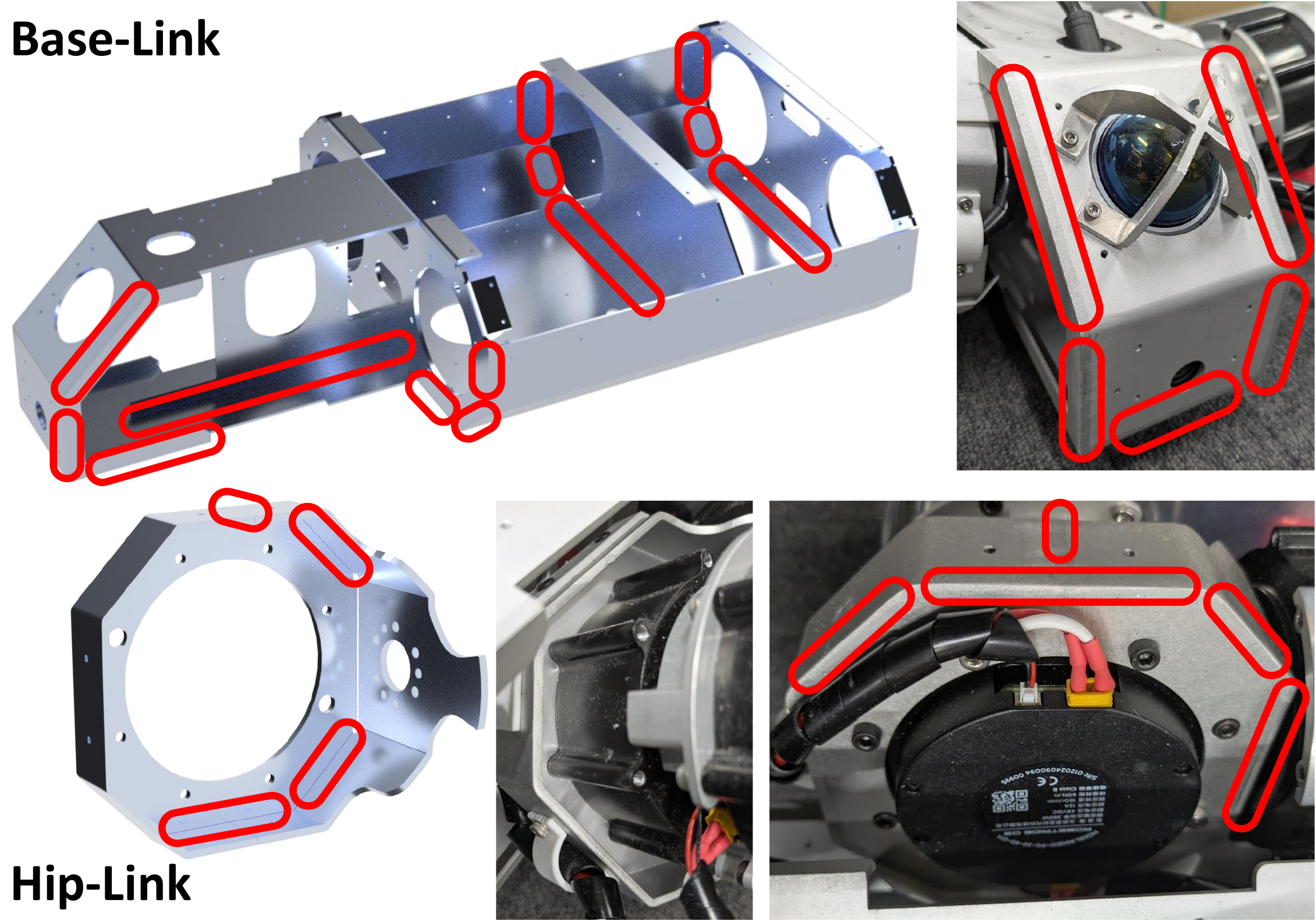}
  \vspace{-1.0ex}
  \caption{Details of sheet metal welding for the Base-Link and Hip-Link. The welded sections are highlighted in red.}
  \label{figure:mevius2-welding}
  \vspace{-1.0ex}
\end{figure}

\begin{figure}[t]
  \centering
  \includegraphics[width=0.8\columnwidth]{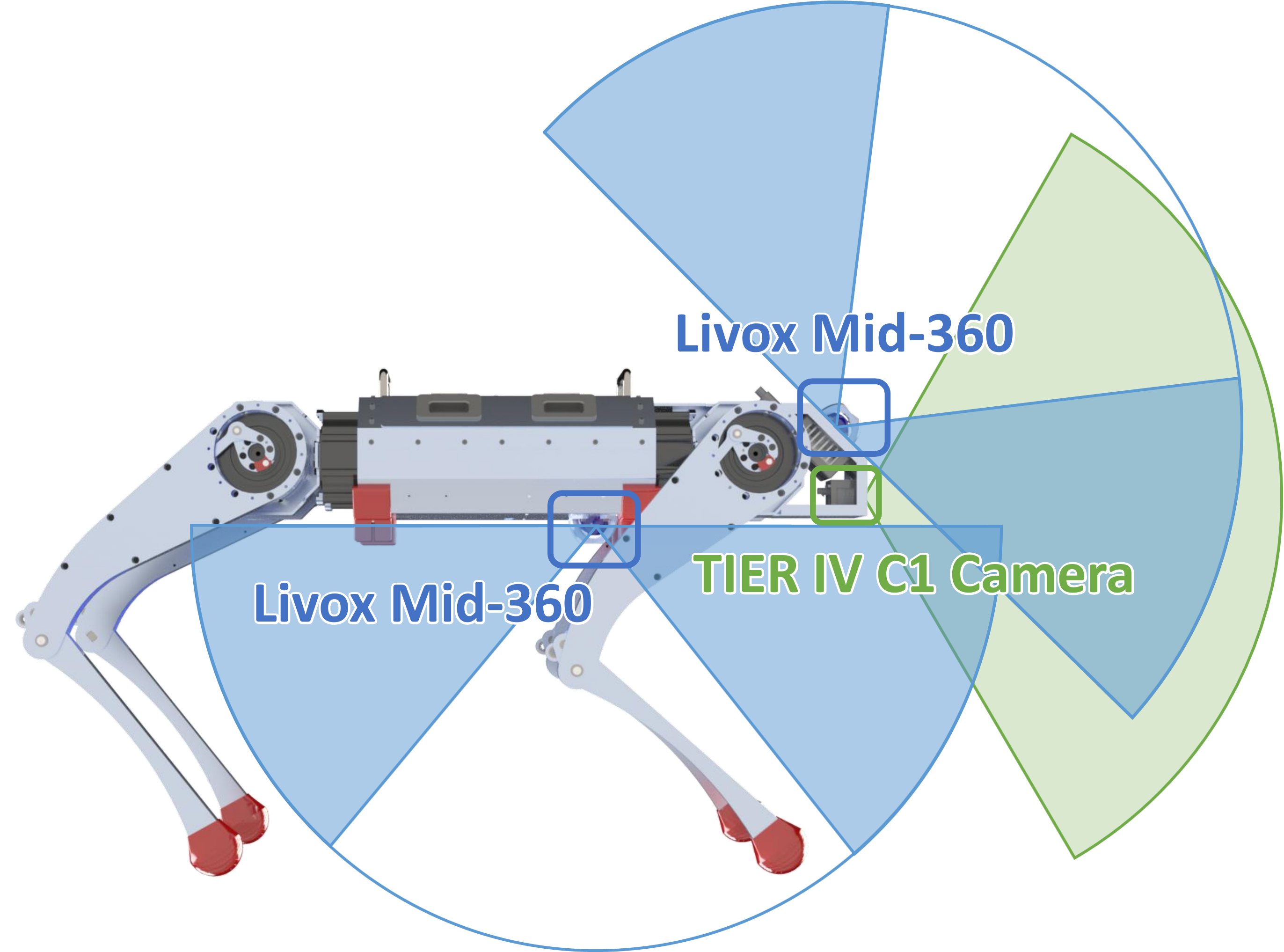}
  \vspace{-1.0ex}
  \caption{The sensor configuration of MEVIUS2.}
  \label{figure:mevius2-sensors}
  \vspace{-3.0ex}
\end{figure}

\section{Design and Configuration of MEVIUS2} \label{sec:mevius2}

\begin{table*}[t]
  \centering
  \caption{Comparison between existing quadruped robots and MEVIUS2}
  \begin{tabular}{l|rrccclrc}
    \multirow{2}{*}{Name} & \multirow{2}{*}{Weight} & \multirow{2}{*}{\shortstack{Leg\\Length\textsuperscript{*1}}} & \multirow{2}{*}{Materials} & \multirow{2}{*}{\shortstack{Open/Closed\\(CAD)}} & \multirow{2}{*}{\shortstack{Off-the-shelf\\(Circuit)}} & \multirow{2}{*}{\shortstack{Maximum\\Torque}} & \multirow{2}{*}{\shortstack{Cost\\(USD)}} & \multirow{2}{*}{Perception} \\
    & & & & & & & & \\ \hline
    Mini Cheetah \cite{katz2019minicheetah}    & 9.0 kg  & 0.20 m & \textbf{Metal}           & Closed        & No           & 17 Nm                     & -    & -                     \\
    ANYMAL \cite{hutter2016anymal}             & 30.0 kg & 0.25 m & \textbf{Metal}           & Closed        & No           & 40 Nm                     & 150k & \textbf{LiDAR/Camera} \\
    Spot \cite{bostondynamics2025spot}         & 33.8 kg & 0.34 m & \textbf{Metal}           & Closed        & No           & 85 Nm\textsuperscript{*2} & 75k  & \textbf{LiDAR/Camera} \\
    Solo-12 \cite{leziart2021solo12}           & 2.5 kg  & 0.16 m & Plastic                  & \textbf{Open} & No           & 2.5 Nm                    & 12k  & -                     \\
    PAWDQ \cite{joonyoung2021pawdq}            & 12.7 kg & 0.22 m & Plastic                  & \textbf{Open} & \textbf{Yes} & 21 Nm                     & 8k   & Camera                \\
    MEVIUS \cite{kawaharazuka2024mevius}       & 15.5 kg & 0.25 m & \textbf{Metal} / POTICON & \textbf{Open} & \textbf{Yes} & 25 Nm                     & 12k  & -                     \\
    MEVIUS2 (this study)                       & 22.9 kg & 0.30 m & \textbf{Metal}           & \textbf{Open} & \textbf{Yes} & 60 Nm                     & 13k  & \textbf{LiDAR/Camera} \\
  \end{tabular}
  \label{table:comparison}
  \begin{flushleft}
  \footnotesize
    \textsuperscript{*1} The average of the link lengths of the thigh and calf links. \\
    \textsuperscript{*2} Because the maximum torque varies depending on the joint angle, we use the average within the joint range of motion.
  \end{flushleft}
  \vspace{-3.0ex}
\end{table*}

\subsection{Design Overview of MEVIUS2} \label{subsec:mevius2-design}
\switchlanguage%
{%
  The design of MEVIUS2 developed in this study is shown in \figref{figure:mevius2-design}.
  MEVIUS2 is composed of 16 metal components, as illustrated in \figref{figure:mevius2-parts}.
  Among these, 11 parts are machined from A7075 aluminum alloy, while the remaining 5 parts are sheet metal components made from A5052.
  Three of the sheet metal components -- Base-Link, Hip-Link, and Lidar-Cover -- are fabricated through sheet metal welding, as detailed in \figref{figure:mevius2-welding}.
  Although most of the structural components are made of metal, the foot tips and body-supporting contact points are 3D printed in TPU to enable soft contact with the environment.
  MEVIUS2 uses 12 Robstride03 motors, each with a continuous torque of 20 Nm and a peak torque of 60 Nm.
  A parallel-link mechanism is employed to transmit the knee joint motion from a proximally located motor, resulting in lighter distal leg segments.

  A key feature of this design is the structure of the Base-Link and the sensor layout.
  The Base-Link consists of five welded sheet metal plates and integrates all mounting points for the legs, circuit components, LiDARs, and RGB camera into a single part.
  This allows for a significant reduction in the number of components while enabling the construction of a large, high-strength body structure.
  All of these metal parts can be automatically quoted and ordered from STEP files using meviy \cite{misumi2024meviy}, MISUMI's machining and fabrication service, making it possible to complete the entire procurement process through e-commerce.
  Notably, this leg configuration uses the fewest number of components among quadrupeds with parallel-link mechanisms \cite{kawaharazuka2024mevius}, enabling a highly simplified assembly process.

  The robot's body is equipped with two LiDAR sensors (Livox Mid-360) and a high dynamic range RGB camera (Tier IV C1 Camera), which together provide detailed environmental perception.
  The perception coverage area of these sensors is illustrated in \figref{figure:mevius2-sensors}.
}%
{%
  本研究で開発したMEVIUS2の設計を\figref{figure:mevius2-design}に示す.
  MEIVUS2は\figref{figure:mevius2-parts}に示すような16の金属部品から構成されている.
  このうち11部品はA7075の金属切削, 他5部品はA5052の板金である.
  板金部品のうち3部品(Base-Link, Hip-Link, Lidar-Cover)は板金溶接により加工されており, それらの詳細は\figref{figure:mevius2-welding}に示されている.
  構造部品は基本的に全て金属であるが, 脚先や胴体を支える部品についてはTPUで3Dプリントされており, 環境との柔らかな接触を可能にしている.
  モータはRobstride03を12個使用しており, その最大連続トルクは20Nm, 最大トルクは60Nmである.
  平行リンク機構により, 膝関節の動きをより近位に配置されたモータから伝達しており, 脚先を軽くしている.

  本設計で特筆すべきはBase-Linkの構成とセンサ配置である.
  Base-Linkは5枚の板金を溶接して構成されており, 脚の取り付け位置, 回路関係部品の取り付け位置, Lidarの取り付け位置, RGBカメラの取り付け位置を全て一部品に備えている.
  これにより, 部品点数を大幅に削減しつつ, 非常に大きな身体構造を高い強度を持って構築することができている.
  これら全ての金属部品はMISUMIの切削・加工サービスであるmeviy \cite{misumi2024meviy}により, STEPファイルから自動的に見積り・発注することが可能であり, 全ての操作をE-Commerceで簡潔させることができる.
  なお, 本脚構成は平行リンクを使う構造の中でも最小の部品点数で構成されており\cite{kawaharazuka2024mevius}, アセンブリを最大限簡易にしている点も特徴である.
  また, 胴体には2つのLidar (Livox Mid-360)と, 高解像度RGBカメラ (Tier IV C1 Camera)を搭載しており, 詳細な環境認識を可能にしている.
  それらの認識可能な範囲は\figref{figure:mevius2-sensors}に示されている.
}%

\begin{figure}[t]
  \centering
  \includegraphics[width=0.9\columnwidth]{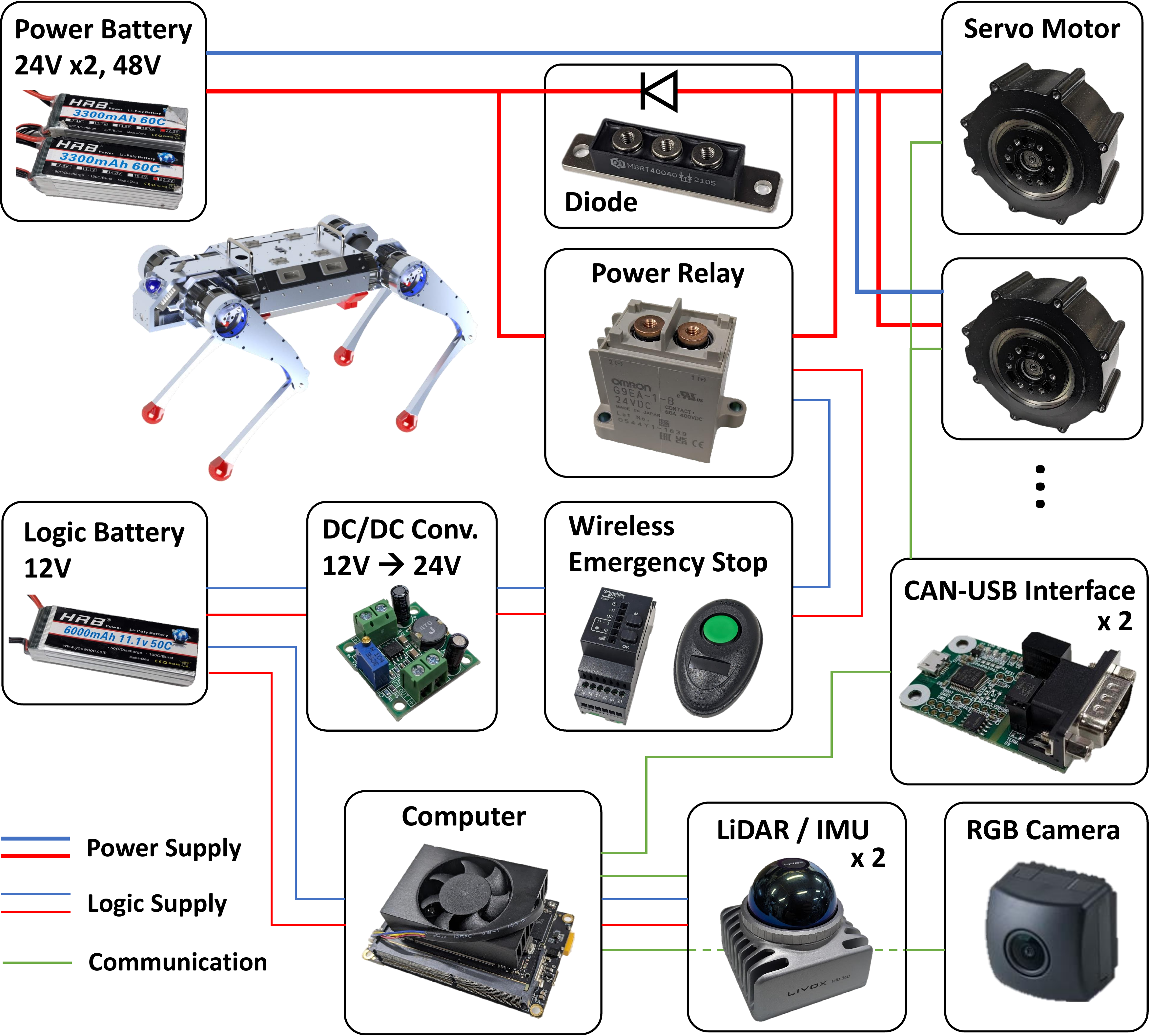}
  \vspace{-1.0ex}
  \caption{Circuit configuration of MEVIUS2: servo motors are connected to the PC via two CAN-USB interfaces, and the system is equipped with a wireless emergency stop, power relay, diode, LiDAR/IMU, and RGB camera.}
  \label{figure:mevius2-circuit}
  \vspace{-3.0ex}
\end{figure}

\begin{figure}[t]
  \centering
  \includegraphics[width=0.9\columnwidth]{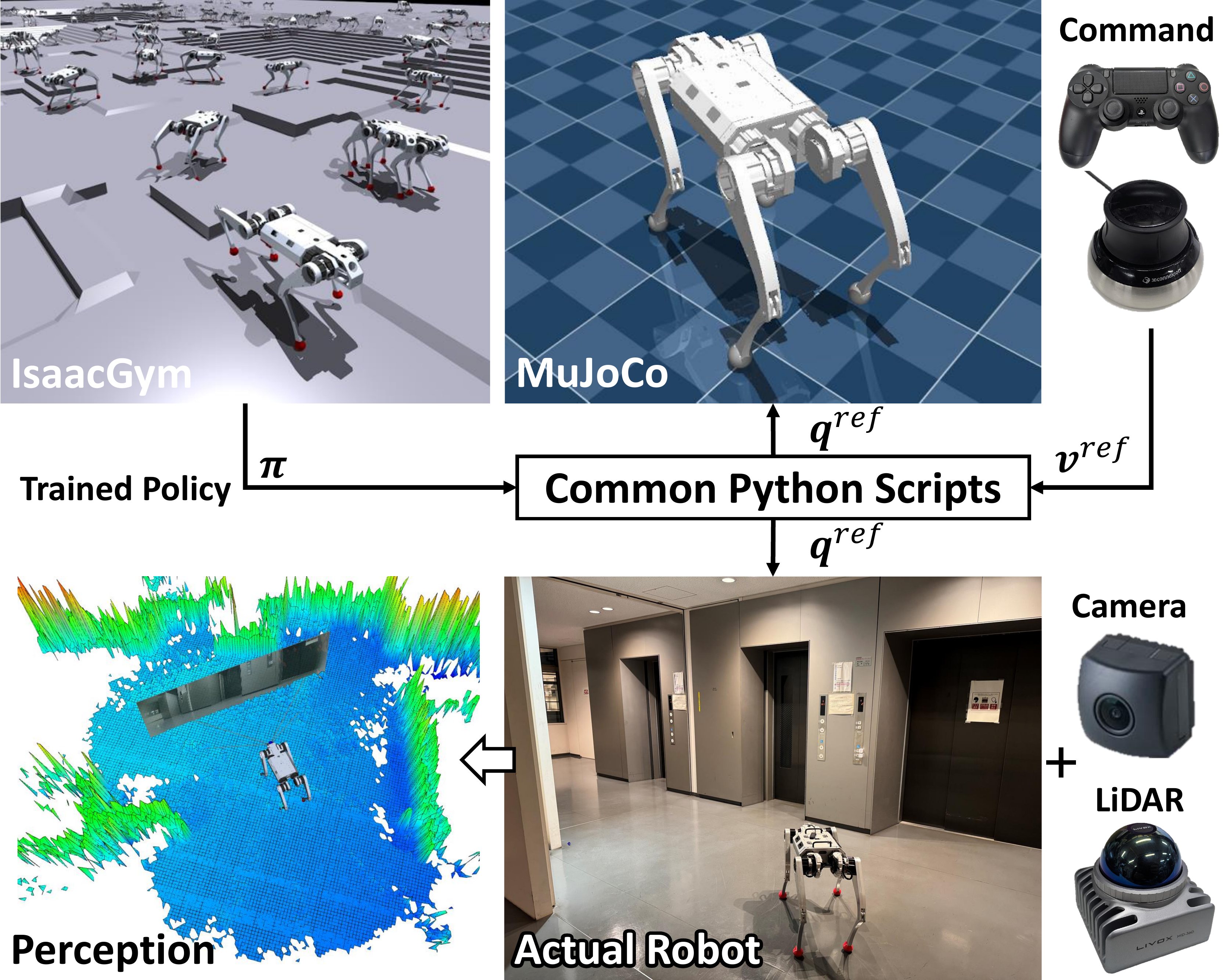}
  \vspace{-1.0ex}
  \caption{Control system of MEVIUS2: policies trained in IsaacGym are verified in MuJoCo and deployed to the actual hardware using common Python scripts.}
  \label{figure:mevius2-control}
  \vspace{-3.0ex}
\end{figure}

\begin{figure*}[t]
  \centering
  \includegraphics[width=1.9\columnwidth]{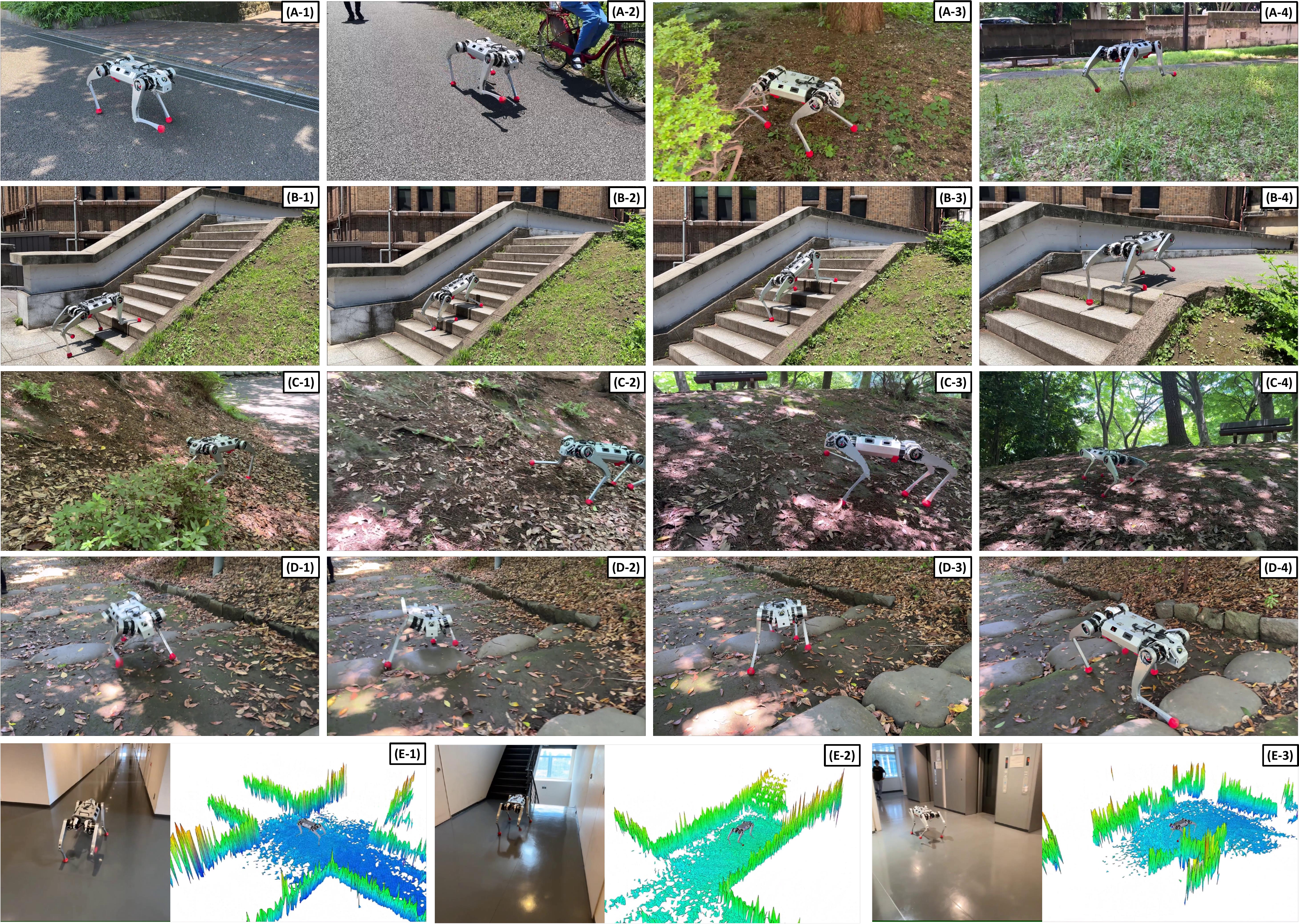}
  \vspace{-1.0ex}
  \caption{Walking experiments in various environments: (A) concrete, grass, and soil; (B) stairs; (C) steep slopes; (D) slippery stairs after rainfall; and (E) indoor settings.}
  \label{figure:outdoor-exp}
  \vspace{-3.0ex}
\end{figure*}

\subsection{Comparison with Existing Quadruped Robots} \label{subsec:comparison}
\switchlanguage%
{%
  A comparison between MEVIUS2 and previously developed quadruped robots is shown in \tabref{table:comparison}.
  Representative commercial quadruped robots include the MIT Mini Cheetah \cite{katz2019minicheetah}, ANYMAL \cite{hutter2016anymal}, and Boston Dynamics Spot \cite{bostondynamics2025spot}.
  These robots differ in size, weight, and maximum torque, but all are constructed from metal and do not have their design or circuit details released as open-source.
  In contrast, Solo-12 \cite{leziart2021solo12}, PAWDQ \cite{joonyoung2021pawdq}, and MEVIUS \cite{kawaharazuka2024mevius} are open-source robots, allowing anyone to build them from scratch.
  Among them, only MEVIUS is constructed from metal -- despite the use of a specialized plastic (POTICON) for some parts -- while the others are primarily made of 3D-printed plastic components.
  MEVIUS2, on the other hand, is a fully open-source quadruped robot made of metal, with size, weight, and maximum torque comparable to ANYMAL and Spot.
  Furthermore, it is equipped with more sensors than other open-source quadruped robots, enabling more practical and detailed environmental perception.
  Its cost is also significantly lower than that of commercial robots of comparable size, making it more accessible to a wider range of researchers.
}%
{%
  MEVIUS2とこれまで開発されてきた四脚ロボットの比較を\tabref{table:comparison}に示す.
  これまで開発されてきた代表的な商業用のロボットとしては, MIT Mini Cheetah \cite{katz2019minicheetah}, ANYMAL \cite{hutter2016anymal}, Boston Dynamics Spot \cite{bostondynamics2025spot}などが挙げられる.
  これらはそれぞれサイズ感や重さ, 最大トルクも異なるが, どれも金属により構成されており, 設計や回路の詳細はオープンソースとして公開されていない.
  これに対して, Solo-12 \cite{leziart2021solo12}やPAWDQ \cite{joonyoung2021pawdq}, MEVIUS \cite{kawaharazuka2024mevius}はオープンソースとして公開されており, 誰でも一から構築することができる.
  このうち, 一部特殊なプラスッチックであるPOTICONが使われているものの, MEVIUSのみが金属により構成されており, その他は3Dプリンタ製のプラスチックにより構成されている.
  一方で, MEVIUS2はANYMALやSpotと同等のサイズ・重量・最大トルクを持つ金属製の四脚ロボットでありながら, 全てのハードウェアがオープンソースとして公開されている.
  また, その他のオープンソース四脚ロボットに比べて多くのセンサを搭載しており, より実用的な環境認識が可能である.
}%

\subsection{Circuit Configuration of MEVIUS2} \label{subsec:mevius2-circuit}
\switchlanguage%
{%
  The circuit configuration of MEVIUS2 is shown in \figref{figure:mevius2-circuit}.
  The system adopts a very simple architecture, where 12 servo motors are controlled via two CAN-USB interfaces.
  For motor power, two 24V, 3300mAh LiPo batteries are connected in series, while a separate 12V, 6000mAh LiPo battery is used as a logic battery for powering the control circuits.
  With this configuration, approximately one hour of continuous operation is possible; however, since there is sufficient space in the payload area, a higher-capacity battery can also be installed.
  The onboard PC is a NVIDIA Jetson, to which two LiDAR sensors and an RGB camera are connected; image and point cloud processing is performed using the GPU on the Jetson.
  Other components include a wireless emergency stop system, power relays, and diodes -- all of which, like the metal structural parts, can be procured entirely via e-commerce platforms.
}%
{%
  MEVIUS2の回路構成を\figref{figure:mevius2-circuit}に示す.
  回路構成は12個のサーボモータ2つのCAN-USBインターフェースから操作する非常にシンプルな構成となっている.
  モータ用のパワーバッテリーは24V 3300mAhのLiPoバッテリーを2つ直列に接続して使用しており, 回路系へのロジックバッテリーは12V 6000mAhのLiPoバッテリーを使用している.
  これらで約1時間の連続稼働が可能であるが, 積載空間には余裕があるため, より大容量のバッテリーを搭載することも可能である.
  PCはNVIDIA Jetsonを用い, Lidar2つと高解像度RGBカメラが接続されており, Jetson上のGPUを用いて画像・点群の処理を行う.
  その他には無線緊急停止装置, パワーリレー, ダイオードなどが接続されており, 金属加工部品と同じく全ての部品をE-Commerceにより調達可能である.
}%

\subsection{Control Architecture of MEVIUS2} \label{subsec:mevius2-control}
\switchlanguage%
{%
  The control architecture of MEVIUS2 is shown in \figref{figure:mevius2-control}.
  MEVIUS2's locomotion is primarily trained through reinforcement learning conducted in IsaacGym \cite{rudin2022leggedgym}.
  The resulting policy is first validated on MuJoCo using a common Python script and then deployed to the actual hardware.
  Details on the reward function and parameters used in the reinforcement learning policy can be found in \href{https://github.com/haraduka/mevius2}{\textcolor{magenta}{github.com/haraduka/mevius2}}.
  For perception, elevation mapping \cite{miki2022elevation} is employed, allowing MEVIUS2 to precisely perceive its surroundings using data from both the LiDAR sensors and the RGB camera.
}%
{%
  MEVIUS2の制御アーキテクチャを\figref{figure:mevius2-control}に示す.
  MEVIUS2の歩行は基本的にIsaacGym上での強化学習に基づき学習されている \cite{rudin2022leggedgym}.
  得られた方策は共通のPythonスクリプトを用いてMuJoCo上で検証され, その後実際のハードウェアにデプロイされる.
  なお, 強化学習の方策に関する報酬やパラメータ等についてはSupplementary Materialsを参照されたい.
  また, perceptionにはelevation mapping \cite{miki2022elevation}を用いており, LiDARとRGBカメラから得られた情報を元に, 周囲の環境を詳細に認識することが可能である.
}%

\section{Experiments} \label{sec:experiment}

\switchlanguage%
{%
  To verify the robot's ability to traverse various types of uneven terrain, walking experiments were conducted in the following environments: (A) concrete, grass, and soil; (B) stairs; (C) steep slopes; (D) slippery stairs after rainfall; and (E) indoor settings.
  These experiments are shown in \figref{figure:outdoor-exp}.
  MEVIUS2 successfully walked not only on diverse terrains including stairs and steep slopes but also managed to maintain walking without falling, even in slippery conditions where it occasionally lost balance.
  In particular, stairs like those in (B), which are commonly used by humans, could not be traversed by the smaller MEVIUS \cite{kawaharazuka2024mevius}; however, MEVIUS2 was able to climb them without any issues, which is a significant improvement.
  Furthermore, during the indoor walking experiments, elevation mapping using LiDAR and a camera was performed, demonstrating that detailed shapes of the surrounding environment could be captured.
}%
{%
  多様な不整地を踏破可能であることを確認するため, (A) コンクリートや芝生, 土, (B) 階段, (C) 急な坂, (D) 雨の後の滑りやすい階段, (E) 室内で歩行実験を行った.
  その様子を\figref{figure:outdoor-exp}に示す.
  階段や急な斜面を含む多様な不整地を歩行できるだけでなく, 滑りやすい状況でも, バランスを崩しながらも倒れること無く最後まで歩行することができた.
  特に, (B)のような一般的に人間が使用する階段はサイズの小さなMEVIUS \cite{kawaharazuka2024mevius}では踏破できなかったが, MEVIUS2では問題なく踏破できた点は大きい.
  また, 室内での歩行実験ではLiDARとカメラを用いたelevation mappingの作成を行い, 周囲の詳細な形状を取得出来ることが分かった.
}%

\section{CONCLUSION} \label{sec:conclusion}
\switchlanguage%
{%
  In this study, we developed MEVIUS2, a practical, metal-based open-source quadruped robot that incorporates sheet metal welding and multimodal perception.
  Compared to 3D-printed parts, metal components are more challenging to fabricate in arbitrary shapes; however, by leveraging sheet metal welding, we were able to construct a large and robust body structure as a single integrated component, significantly reducing the number of individual parts.
  Equipped with LiDARs and a high dynamic range camera, MEVIUS2 can acquire detailed environmental information, achieving a more practical configuration in terms of both scale and sensing compared to existing open-source quadruped robots.
  With the ability to traverse various types of rough terrain through reinforcement learning and a high degree of customizability due to the full release of all hardware and software, we anticipate that MEVIUS2 will serve as a foundation for a wide range of future research.
}%
{%
  本研究では, 板金溶接とマルチモーダル認識を用いた実用的なサイズの金属製オープンソース四脚ロボットMEVIUS2を開発した.
  3Dプリンタに比べて自由形状の難しい金属部品を, 板金溶接を駆使することで, 非常に大きな身体構造を強度を保ちつつ一部品として構築し, 部品点数を大幅に削減することができる.
  またLiDARと高解像度カメラにより環境情報を詳細に取得可能であり, サイズやセンサ装備ともに, これまでのオープンソース四脚ロボットに比べ実用的な構成を可能にしている.
  強化学習により多様な不整地を踏破可能であり, 全ての部品やソフトウェアが公開されているというカスタマイズ性の高さから, 今後様々な研究に発展することを期待する.
}%

{
  \bibliographystyle{IEEEtran}
  \bibliography{main}
}

\newpage
\section*{Description of Supplementary Materials} \label{sec:appendix}

\subsection{Overview}
\switchlanguage%
{%
  The Supplementary Materials for this study include the following files:
  \begin{itemize}
    \item \textbf{PartsList.xlsx}: A list of all components used in MEVIUS2.
    \item \textbf{MEVIUS2-STEP.tar.gz}: STEP files for each individual MEVIUS2 component.
    \item \textbf{MEVIUS2-Assembly.STEP}: The full assembly file of MEVIUS2.
    \item \textbf{LeggedGym.tar.gz}: The reinforcement learning environment for MEVIUS2.
    \item \textbf{Software.tar.gz}: The software package for MEVIUS2.
    \item \textbf{Video.mp4}: The video of experiments.
  \end{itemize}
}%
{%
  本研究のSupplementary Materialsに含まれるファイルは以下の通りである.
  \begin{itemize}
    \item PartsList.xlsx: MEVIUS2の部品リスト.
    \item MEVIUS2-STEP.tar.gz: MEVIUS2の各部品のSTEPファイル.
    \item MEVIUS2-Assembly.STEP: MEVIUS2のアセンブリファイル
    \item LeggedGym.tar.gz : MEVIUS2の強化学習環境.
    \item Software.tar.gz: MEVIUS2のソフトウェア.
    \item Video.mp4: MEVIUS2の実験動画.
  \end{itemize}
}%

\subsection{Hardware}
\switchlanguage%
{%
  This section provides information related to the hardware, with a focus on key points regarding the parts list.
  Currently, the total cost of MEVIUS2's metal components is approximately 6,150 USD, and the full system -- including motors, sensors, and other components -- amounts to 12,905 USD.
  The parts list includes details such as part names, materials, quantities, and prices, but the most critical information is the part numbers used for ordering via meviy.
  All metal components used in this study can be ordered simply by specifying their part numbers.
  Of course, it is also possible to place orders by uploading the corresponding STEP files.
  The quoted price is based on a 20-day delivery time; however, a shorter lead time of approximately three days is also possible at the earliest.
  All mechanical parts can be ordered through MISUMI and meviy, and for electronic components, individual URLs are provided in the list.
}%
{%
  ここではハードウェアに関する情報を提供する.
  特に, 部品リストについて主要なポイントを解説する.
  まず, 現状MEVIUS2の金属部品は全部で6150ドル, モータやセンサなどを全て入れると12905ドルとなっている.
  部品リストには名前, 素材, 個数, 価格などが書かれているが, 特に重要なのはmeviyで発注する部品の型番である.
  本研究の金属部品は全て, 型番を指定するのみで発注することができる.
  もちろん, STEPファイルをアップロードして発注しても良い.
  提示した価格が20日納期の場合であるが, 最短で3日程度の納期も可能である.
  全ての機械部品がMISUMIとmeviyによって発注可能であり, 回路部品については個別にURLが記載されている.
}%

\subsection{Software}
\switchlanguage%
{%
  This section provides information related to the software.
  The reinforcement learning environment is based on LeggedGym and includes the URDF model of MEVIUS2.
  The reward functions and parameters used are consistent with those commonly applied to quadruped robots.
  The control software includes the Python scripts necessary for operating MEVIUS2, covering both MuJoCo-based Sim-to-Sim simulation and real-world Sim-to-Real deployment.
  It also includes launch programs for the LiDARs and camera, which can be visualized using ROS and Rviz.
  Accordingly, the package also contains the MuJoCo XML model files and URDF files for Rviz visualization.
}%
{%
  ここではソフトウェアに関する情報を提供する.
  強化学習環境はLeggedGymをベースにしており, MEVIUS2のURDFモデルが含まれている.
  報酬やパラメータは一般的な四脚ロボットのものを使用している.
  制御ソフトウェアはMEVIUS2の制御に必要なPythonスクリプトが含まれており, Sim-to-Sim用のMuJoCoとSim-to-Real用の実機制御用のコードが含まれている.
  また, LiDARやカメラの立ち上げプログラムも含まれており, それらはROSとRvizを用いて描画可能である.
  そのため, MuJoCo用のXMLモデルファイル, Rviz用のURDFモデルファイルなども含まれている.
}%

\subsection{Limitations}
\switchlanguage%
{%
  Several limitations of MEVIUS2 are discussed below.
  First, the current MEVIUS2 does not feature a waterproof or dustproof design, and therefore special care is required when operating it outdoors.
  While this is a critically important issue for future development, realizing a waterproof and dustproof structure in a form that can be released as open source is extremely challenging and remains an open problem.
  Second, although the welded metal structure can be fabricated relatively easily by using manufacturing services such as meviy or conventional machining vendors, building it from scratch as an individual is more difficult compared to standard metal machining processes.
  As a means to achieve a large metal-based quadruped robot at low cost with a small number of parts, it is necessary to explore alternatives to sheet metal welding in future work.
  Third, while this study implements environmental perception using LiDARs and cameras, achieving truly practical environmental perception will likely require the integration of a wider variety of sensors, such as RTK-GNSS, thermal cameras, and foot contact sensors.
  Future work should therefore focus on incorporating additional sensors and developing more advanced perception methods that leverage them.
  Finally, safety remains an important issue.
  MEVIUS2 has not undergone formal risk assessment, and thus must be operated with sufficient caution.
  Safety-related challenges, such as pinch hazards and behavior during emergency stops, need to be addressed in future work.
}%
{%
  MEVIUS2に関するいくつかの限界について述べる.
  まず, 現状のMEVIUS2は防水・防塵構造にはなっておらず, 屋外での使用には注意が必要である.
  これは今後非常に重要な点であるが, オープンソースとして公開できる形で防水・防塵構造を実現することは非常に難しく, 今後の課題である.
  次に, 板金溶接による構造は, meviyのような加工サービスや機械加工業者を利用すれば簡単に制作可能であるが, 個人で一から制作することは一般的な金属切削に比べると難しい.
  安く, 少部品で, 大きな金属筐体を実現するための手段として, 今後板金溶接以外にも選択肢がないか検討していく必要がある.
  次に, 本研究ではLiDARとカメラを用いた環境認識を実現しているが, 本当に実用的な環境認識を行うためには, RTK-GNSSや熱画像カメラ, 脚先の接触センサなど, さらに多様なセンサを搭載する必要がある可能性が高い.
  今後, さらなるセンサの搭載や, それらを用いた高度な環境認識手法の実装を進めていく必要がある.
  最後に, 安全性の課題がある.
  MEVIUS2にはリスクアセスメントなどは施されていないため, 使用にあたっては十分な注意が必要である.
  手の挟み込みや緊急停止時の挙動など, 安全性に関する課題を今後解決していく必要がある.
}%

\end{document}